\documentclass{article}
\usepackage{multirow}
\usepackage{amsmath,graphicx,mlspconf}
\usepackage{xcolor}
\usepackage{svg}
\name{Xiaoyu Luo, Qiongxiu Li\thanks{This work is partially supported by the EU ChipsJU and the Innovation Fund Denmark through the project CLEVER (no. 101097560).}}
\address{Department of Electronic Systems, Aalborg University, Copenhagen, Denmark}

\begin{document}

\title{DeMem: Privacy-Enhanced Robust Adversarial Learning via De-Memorization}

\author{
\IEEEauthorblockA{
}
}

\maketitle
\begin{abstract}
Balancing adversarial robustness and privacy in machine learning models is essential yet particularly challenging. Previous studies have shown that enhancing adversarial robustness through adversarial training often increases vulnerability to privacy attacks, revealing a fundamental tension between these two attributes. How to reduce privacy leakage in robust models has received little attention. Existing approaches like differential privacy (DP) remain ineffective in adversarial training, as it can severely degrade performance (e.g., test accuracy dropping from 50.87\% to 20.35\% on CIFAR-100).  Through a detailed analysis, we find that such limitation arises because DP treats all training samples uniformly, failing to account for the diverse privacy risks and generalization abilities of individual samples. This oversight disproportionately impacts relatively low-risk samples which are often the typical samples that are essential for model robustness, leading to undesired performance degradation.
To address this limitation, we propose DeMem, a novel method that selectively targets high-risk samples to achieve a better balance between privacy protection and adversarial robustness. By leveraging sample-wise granularity, DeMem minimizes privacy leakage without compromising robustness.
Extensive evaluations across multiple datasets and adversarial training methods demonstrate that DeMem significantly reduces privacy leakage while maintaining robust performance against both natural and adversarial samples. (e.g., implementing DeMem with PGD-AT on CIFAR-10 reduces privacy leakage risk by 8\% without compromising both nature and robust accuracy). These results highlight DeMem’s versatility, effectiveness, and broad applicability in enhancing the trustworthy attributes of machine learning models.

\end{abstract}


\section{Introduction}
\label{sec:intro}
As machine learning models, particularly deep neural networks (DNNs), become integrated into daily life, ensuring their trustworthiness—including robustness, privacy, and fairness \cite{9933776,chang2021privacy,li2024adbm,li2024language,li2024faster}—is crucial. Without these qualities, deploying models in sensitive areas like healthcare, finance, and autonomous systems risks biases, privacy breaches, and malicious manipulation. Thus, ensuring reliability and integrity is essential for the safe and responsible use of machine learning.

Balancing these attributes, however, is a major challenge. Prior work \cite{song2019privacy,li2024privacy} shows that adversarial training, while enhancing robustness, increases vulnerability to membership inference attacks (MIA) \cite{shokri2017membership}. Conversely, applying Differential Privacy (DP) can significantly weaken robustness against adversarial examples \cite{tursynbek2020robustness}. This reveals the complex trade-offs in designing deep neural networks that balance these attributes. However, enhancing privacy in robust models is underexplored, as most studies optimize robustness or privacy independently, often overlooking their interplay in adversarial settings.

To mitigate privacy risks in adversarial learning, a straightforward approach is to apply DP, which offers strong privacy guarantees. However, combining DP with adversarial training increases the learning challenge, causing significant performance degradation \cite{hayes2022learning} – as seen in CIFAR-10, where natural test accuracy of PGD-AT can drop from 81.34\% to 38.46\% by applying DP-SGD \cite{abadi2016deep}(see Table \ref{table:performance} for details). The performance decline likely stems from noise perturbations in DP increasing the loss landscape's curvature, leading to poorer generalization \cite{hayes2022learning}. Although DP protects privacy, the resulting performance drop undermines the robustness adversarial training seeks to achieve. Balancing privacy without sacrificing robustness thus remains an open challenge.

To address this challenge, we first conduct an in-depth analysis of why applying DP significantly degrades model performance. Our analysis takes a novel approach by examining individual sample memorization scores \cite{feldman2020does, feldman2020neural}, which represent each sample's privacy leakage risk. A high memorization score indicates that a sample significantly influences the model’s output, implying a higher privacy risk if used during training.  Through detailed investigation, we reveal that DP disproportionately impacts samples with low privacy risks—those that are crucial for maintaining overall model performance. This explains the observed decline in robustness. 

Building on these findings,  we argue that restricting samples with high privacy risks can effectively enhance privacy without significantly compromising model performance. 
To address this limitation, we propose DeMem, a novel approach that selectively targets high-risk samples to mitigate privacy leakage without compromising performance. DeMem allowing to focus on sample-wise granularity while preserving the utility of the model.
Our main contributions are summarized as follows: 

1) We conduct a nuanced analysis to attribute why DP degrades model performance, focusing on the individual sample's memorization score to uncover the impact on privacy risk and robustness.

2) We reveal that DP disproportionately harms samples with low privacy risks, which are essential for preserving model robustness.

3) We propose DeMem, a method that selectively restricts high-risk samples to enhance privacy while minimizing performance degradation.

DeMem can be seamlessly integrated into various adversarial training techniques. 
Extensive experiments on various adversarial training methods, including PGD-AT \cite{madry2017towards} and TRADES \cite{zhang2019theoretically}, and datasets such as CIFAR-10 and CIFAR-100, demonstrate the effectiveness of DeMem in enhancing privacy while preserving robustness against both natural and adversarial samples. Our approach achieves a better balance between privacy and robustness in machine learning.

\section{Preliminaries}
In this section, we review necessary fundamentals that are essential for understanding the rest of the paper.

\subsection{Adversarial robustness}
Robustness is essential for the reliable deployment of machine learning models, particularly in safety-critical domains like autonomous driving. However, adversarial examples \cite{goodfellow2014explaining} pose a major challenge. For example, a small perturbation can cause a stop sign to be misclassified as a speed limit 45 sign, threatening the safety of autonomous vehicles \cite{eykholt2018robust}. Adversarial training is among the most effective methods for enhancing model robustness \cite{madry2017towards, zhang2019theoretically, wu2020adversarial}. Approaches such as PGD-AT \cite{madry2017towards} and TRADES \cite{zhang2019theoretically} improve robustness by incorporating adversarial examples during training to minimize adversarial risk.
\subsection{Differential privacy}
DP \cite{dwork2006differential} provides a rigorous definition of privacy. Where \( M \) denotes any algorithm, It ensures that the output of \( M \) does not significantly differ where \( D \) and \( D' \) denote any two neighboring datasets with one datapoint difference.  \( S \subseteq \text{Range}(M) \) is any event, \( \epsilon \) is a non-negative parameter that quantifies the privacy loss, and \( \delta \) is a small positive term that allows for a slight probability of failure. Formally, DP is defined as follows:

\[
P[M(D) \in S] \leq e^{\epsilon}P[M(D') \in S] + \delta
\]

This definition ensures that the presence or absence of a single data point has a limited effect on the algorithm's output distribution, thereby protecting individual privacy.

To implement DP, Differentially Private Stochastic Gradient Descent (DP-SGD) \cite{abadi2016deep} provides a solution that protects data privacy by clipping the gradients and adding Gaussian noise during each update step. 

\subsection{Memorization score}
The memorization score \cite{feldman2020does, feldman2020neural} quantifies how sensitive a model's output is to the inclusion or exclusion of a specific data point.  Let \(\mathcal{A}\) denote a learning algorithm, \( D_{\text{tr}} \) denote a training dataset. Formally, for a data point \( i \subseteq D_{\text{tr}} \), its (label) memorization score is defined as:
\begin{align}\label{eq:mem}
\text{mem}(\mathcal{A}, D_{\text{tr}}, i) 
&= \Pr_{h \leftarrow \mathcal{A}(D_{\text{tr}})} \left[ h(x_i) = y_i \right]  \nonumber \\
&\quad - \Pr_{h \leftarrow \mathcal{A}(D_{\text{tr}} \setminus \{i\})} \left[ h(x_i) = y_i \right],
\end{align}

where \(D_{\text{tr}} \setminus \{i\} \) denotes the dataset \(D_{\text{tr}}\) with the sample $i$ being removed.

The memorization score thus indicates the potential privacy risk of a sample, based on the model's sensitivity to its inclusion in the training set. Empirical evidence supports this, showing that samples with high memorization scores are often atypical or challenging, while those with low memorization scores tend to be more typical or easier examples across various datasets \cite{feldman2020neural}, consistent with human intuition.

\subsection{Membership inference attack}
MIA \cite{shokri2017membership} is a common method for quantifying privacy leakage by determining whether a specific data point was part of a model’s training dataset. Such membership information can reveal highly sensitive personal details, such as health conditions \cite{alavijeh2019quality}, and can serve as the basis for more advanced privacy attacks \cite{carlini2022membership}.  Many MIAs have been proposed for example confidence-based attacks \cite{salem2018ml}, loss attacks \cite{yeom2018privacy}, and modified entropy attacks \cite{song2021systematic}. Regarding the evaluation metrics of MIAs, it has been argued in many works \cite{rezaei2021difficulty, carlini2022membership, hintersdorftrust}  that a MIA should achieve high true positive rates (TPR) while maintaining as low a false positive rate (FPR) as possible. In other words, the attack should accurately infer the membership status of each data point, ensuring that the inferences are correct and reliable. Achieving a high TPR with a correspondingly low FPR is crucial, as it reflects the effectiveness of the attack in correctly identifying members without erroneously classifying non-members as members \cite{carlini2022membership}. This balance is essential to accurately quantify the privacy leakage and understand the vulnerabilities of the model with respect to MIAs. 

Therefore, in this paper, we utilize LiRA \cite{carlini2022membership} as the primary tool to assess privacy leakage in trained models. LiRA represents the state-of-the-art (SOTA) in MIA and is closely aligned with the concept of memorization. Its performance surpasses that of earlier MIA methods \cite{carlini2022membership,li2024privacy}, making it the ideal choice for our analysis.

For evaluation, we focus on measuring the TPR at a low FPR, which is critical for accurately identifying membership without misclassifying non-members as members \cite{rezaei2021difficulty, carlini2022membership, hintersdorftrust}, rather than relying on traditional metrics like overall balanced accuracy.

\section{The proposed approach}
We first review related work on privacy and adversarial robustness, highlighting the challenge of balancing the two. We then analyze individual samples to explain why DP can fail, followed by a detailed presentation of our proposed approach.

\subsection{Tension between Privacy and Adversarial Robustness}

Although no theoretical proof has established an inevitable trade-off between privacy and adversarial robustness, numerous empirical studies have observed its existence \cite{song2019privacy, li2024privacy, tursynbek2020robustness}. From the perspective of enhanced privacy, models trained using DP-SGD in the presence of adversarial examples consistently exhibit a gap in adversarial robustness performance when compared to models trained with standard SGD \cite{tursynbek2020robustness}. Conversely, applying adversarial training tends to make the model more vulnerable to privacy attacks \cite{song2019privacy, li2024privacy}. These findings suggest that there may be an intrinsic and irreconcilable trade-off between the objectives of privacy and adversarial robustness.
Moreover, current adversarial training methods rarely prioritize privacy protection, focusing mainly on improving defenses against adversarial attacks. As a result, robust models can unintentionally expose sensitive information, creating additional risks in privacy-sensitive applications. 

One typical way to enhance privacy is to apply DP, it works by adding noise to the sensitive information for perturbing it from revealing.  However, previous research has shown that the magnitude of noise perturbations used in DP can increase the curvature of the loss landscape, ultimately leading to degraded generalization performance \cite{hayes2022learning}. This further intensifies the difficulty of achieving a balance between privacy and robustness.

As an example, we apply DP in PGD-AT \cite{hayes2022learning} as a privacy-preserving baseline. As shown in Table \ref{table:performance}, even adding a small amount of DP noise during PGD-AT training significantly hinders model convergence, resulting in substantial drops in both natural and robust accuracy. This empirical evidence clearly demonstrates that directly combining DP with adversarial training is not a viable approach, as the performance degradation renders the robust model practically unusable. The severe decline in both natural and robust accuracy further validates the hypothesis of an inherent conflict between ensuring privacy and maintaining adversarial robustness. Thus, it becomes evident that more sophisticated methods are required to address this trade-off effectively, without compromising either privacy or robustness beyond acceptable limits.


\subsection{Analysis via memorization scores}
\label{sec:pivm}
We now analyze the reasons behind the significant drop in model performance caused by DP. Specifically, we explore how the privacy and performance of samples with varying memorization scores are affected after applying DP. 

\begin{figure}[htb]
\centering
\includegraphics[width=6.0cm]{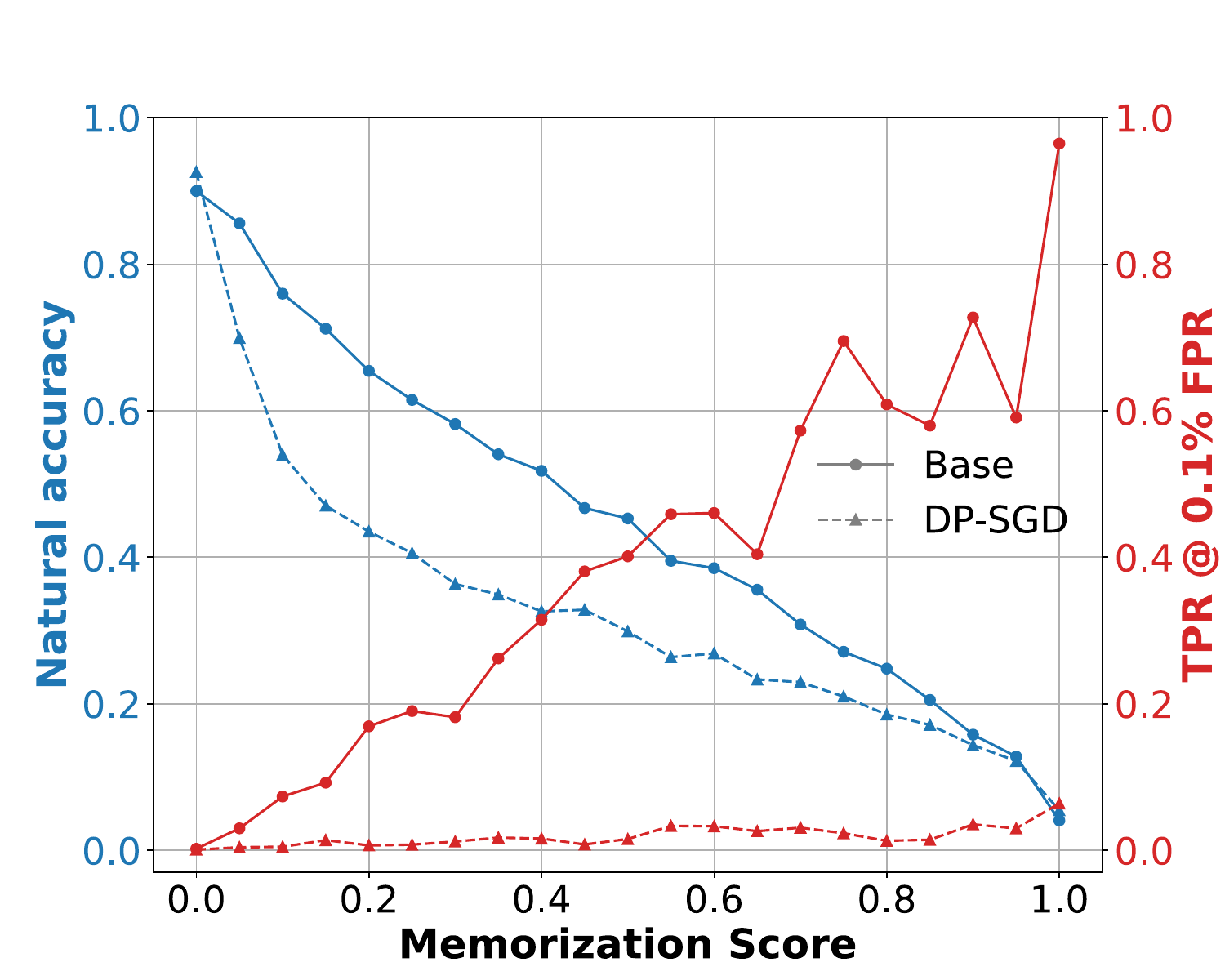}
\caption{The privacy leakage (TPR of the corresponding samples) and test accuracy between baseline and DP model on CIFAR-100, DP enhanced by DP-SGD with a noise multiplier of 0.05 and a max gradient norm of 10.}
\label{fig:privacy_leakage}
\end{figure}

In Figure \ref{fig:privacy_leakage}, we compare the membership privacy leakage and test accuracy of samples with varying memorization scores across two models trained on the CIFAR-100 dataset: one trained with DP-SGD and a baseline model without DP-SGD. In comparison to the baseline, the overall performance of DP-SGD dropped from 70.14\% to 54.13\%, illustrating the trade-off between privacy protection and model accuracy. To ensure an unbiased comparison, we used the memorization scores generated in prior work \cite{feldman2020neural}, which involved training CIFAR-100 on 4,000 ResNet-50 models. Given that most sample memorization values are concentrated at 0, we assign 0 to a separate bin. The remaining range from 0 to 1 is evenly divided into 21 bins with open and closed intervals. 

We calculate the attack success rate of LiRA in each bin and use the TPR at 0.1\% FPR as the privacy leakage risk evaluation metric. It can be observed that the base model exhibits a trend where a higher memorization score corresponds to higher privacy leakage. The models trained with DP-SGD, on the other hand, demonstrate superior privacy protection by limiting the attack success rate for all samples, resulting in less privacy leakage across all memorization scores.  Additionally, the accuracy for samples with lower memorization scores is more notable than the samples with high memorization scores. This suggests that a significant portion of the performance loss arises from samples with lower memorization scores, which originally posed a lower risk of privacy leakage.

This observation highlights the uneven impact of differential privacy across samples. Thus, it is crucial to analyze privacy leakage and performance degradation from a memorization perspective. This helps clarify how privacy mechanisms affect different samples and identify ways to minimize performance loss without compromising privacy.

\subsection{DeMem: DeMemorization}
The performance drop in DP models mainly stems from samples with memorization scores above zero, which pose higher privacy risks. To address this, we selectively target high-memorization samples to preserve overall performance. We propose DeMem, a dememorization-based method that limits the model’s retention of specific samples while minimizing performance impact.

A key challenge is the lack of prior knowledge of memorization scores in practice. As shown in Eq. \eqref{eq:mem}, calculating memorization is computationally expensive, requiring $n+1$ models for $n$ samples in a strict leave-one-out setup, making it impractical for large datasets. Feldman et al. \cite{feldman2020neural} addressed this by precomputing memorization scores with 4,000 ResNet-50 models and sampling tricks.

Given the high computation cost of exact memorization scores, we seek more efficient proxies. Noting that high-memorization samples are often hard or atypical examples with larger losses, we examine the Spearman correlation between loss and memorization scores. As shown in Table \ref{table:spearman_loss_mem}, strong positive correlations (0.501 on training and 0.784 on test) confirm that loss is a reliable indicator of memorization.

\begin{table}[htb]
\renewcommand{\arraystretch}{1.2}
\centering
\caption{Spearman Correlation Coefficient Between Loss and Memorization Score on Training and Test Data Under baseline model on CIFAR-100.}
\begin{tabular}{c|c} 
\textbf{Dataset}        & \textbf{Correlation Coefficient} \\ 
\hline
Training Data           & 0.501                                   \\  
Test Data               & 0.784                                    \\  
\end{tabular}
\label{table:spearman_loss_mem}
\end{table}
Notably, the correlation on the training set (member) is reduced compared to the test set (non-member), reflecting the model's increased fit to the training data. This reduction in correlation suggests that the training process itself increases the model's reliance on memorization for high-loss samples, which are often associated with higher privacy risks.

To address this issue, we propose the following DeMem approach, leveraging the observed relationship between loss and memorization as a basis for targeted interventions. Let $\mathcal{D}$ be a dataset, and $\mathcal{B} \subseteq \mathcal{D}$ be a mini-batch of size $N$ used during the standard training process. For each sample $x_i \in \mathcal{B}$, the corresponding loss is denoted as $\ell(x_i, \theta)$, where $\theta$ represents the model parameters.

We first define a Sample-wise Dememorization Penalty as the variance of the losses computed over the sampled subset $\mathcal{S}$, i.e., 

\[
\Psi(\mathcal{B}) = \frac{1}{N} \sum_{i=1}^{N} \left(\ell(x_i, \theta) - \frac{1}{N} \sum_{j=1}^{N} \ell(x_j, \theta)\right)^2
\]

To incorporate the Sample-wise Dememorization Penalty into the overall loss, we multiply the penalty by a dememorization parameter $\lambda$ and add it to the total loss function $\mathcal{L}(\theta)$:

\[
\mathcal{L}_{\text{total}}(\theta) = \mathcal{L}(\theta) + \lambda \cdot \Psi(\mathcal{B})
\]

Where $\mathcal{L}(\theta)$ is the original loss function without DeMem regularization. $\lambda$ is a hyperparameter controlling the strength of the dememorization penalty.

This observation of Table \ref{table:spearman_loss_mem} also supports our proposed approach. By applying additional penalties to high-loss samples during training, we can reduce the model's memorization of these high-risk data points, effectively mitigating privacy leakage. Thus, using loss as a proxy for memorization not only simplifies computation but also aligns with our objective of improving privacy protection through targeted dememorization.

\section{Experimental results}

\subsection{Setup}
Experiments were conducted on 8 NVIDIA 4090 GPUs using PyTorch \cite{pytorch}.

\textbf{Datasets.} Following previous work \cite{li2024privacy,carlini2022membership}, we employed ResNet-50 \cite{he2016deep} and used the CIFAR-100, CIFAR-10 datasets for MIA and performance evaluations. 

\textbf{MIA setting.} Following LiRA's setting, for each adversarial training method, a total of 128 models were trained. Each used \(\sim\)30,000 training samples (members) and was evaluated on the remaining \(\sim\)30,000 test samples (non-members). Each data point appeared in 64 IN models and was excluded from 64 OUT models for balanced evaluation.

\textbf{DP-SGD setting.} We utilized the Opacus framework \cite{opacus}, which provides DP-SGD with a noise multiplier of 0.05 and a max gradient norm of 10.

\subsection{Results}
All results in this section represent the mean values of 10 models randomly selected from a set of 128 models. 

\textbf{Effectiveness of the proposed DeMem}: 
In Table \ref{table:performance}, the "Base" refers to PGD-AT with DP-SGD configured identically to Table \ref{table:performance}. For both CIFAR-10 and CIFAR-100, the "Base" method demonstrates a notable compromise in performance. In contrast, our proposed method, DeMem, effectively preserves performance.
\begin{table}[htb!]
\renewcommand{\arraystretch}{1.2}
\caption{Model performance comparison on CIFAR-100 and CIFAR-10 for two adversarial training methods, with and without the proposed DeMem.}
\centering
\resizebox{\linewidth}{!}{%
\begin{tabular}{c|c|cc}
\textbf{Dataset}   & \textbf{Method}         & \textbf{Natural Acc. (\%)} & \textbf{Robust Acc. (\%)} \\ \hline
\multirow{4}{*}{CIFAR-100} & Base            & 31.27 ± 0.26        & 13.19 ± 0.16          \\ \cline{2-4}
                           & PGD-AT            & 50.87 ± 0.22        & 15.78 ± 0.11          \\  
                           & PGD-AT + \textbf{DeMem}    & 50.51 ± 0.25        & 15.63 ± 0.14          \\  
                           & TRADES         & 49.37 ± 0.25        & 18.93 ± 0.19          \\ 
                           & TRADES + \textbf{DeMem} & 49.33 ± 0.42        & 16.77 ± 0.31          \\ \hline
\multirow{4}{*}{CIFAR-10}  & Base            & 38.64 ± 0.27        & 23.94 ± 0.26          \\ \cline{2-4}
                           & PGD-AT            & 81.34 ± 0.13        & 35.96 ± 0.21          \\ 
                           & PGD-AT + \textbf{DeMem}    & 80.10 ± 0.32        & 35.03 ± 0.59          \\ 
                           & TRADES         & 80.76 ± 0.15        & 44.36 ± 0.19          \\  
                           & TRADES + \textbf{DeMem} & 80.78 ± 0.45        & 42.86 ± 0.59          \\ 
\end{tabular}%
}
\label{table:performance}
\end{table}

As shown in Tables \ref{table:privacy}, we can observe that our dememorization method can notably reduce the model's privacy leakage under a strict FPR limitation across various datasets, adversarial training method and model structures, with minimal to no natural and robustness performance loss. For instance, the TPR at 0.1\% FPR decreased from 20.29\% to 12.39\% when comparing PGD-AT to PGD-AT + DeMem on CIFAR-10, while the robustness accuracy showed only a minor drop of 0.93\%.
\begin{table}[htb!]
\renewcommand{\arraystretch}{1.2}
\caption{Privacy leakage reduction comparison on CIFAR-100 and CIFAR-10 for two adversarial training methods, with and without the proposed DeMem. The metrics are TPR at different FPR thresholds.}
\centering
\resizebox{\linewidth}{!}{%
\begin{tabular}{c|c|cc}
\textbf{Dataset}   & \textbf{Method}         & \textbf{TPR @ 0.1\% FPR} & \textbf{TPR @ 0.001\% FPR} \\ \hline
\multirow{4}{*}{CIFAR-100} & PGD-AT            & 67.23 ± 0.71 & 44.53 ± 3.72   \\  
                           & PGD-AT + \textbf{DeMem}    & \textbf{64.65 ± 0.69} & \textbf{40.85 ± 4.21}   \\  
                           & TRADES         & 52.60 ± 1.38 & 31.03 ± 3.91   \\ 
                           & TRADES + \textbf{DeMem} & \textbf{47.80 ± 0.88} & \textbf{24.81 ± 4.83}   \\ \hline
\multirow{4}{*}{CIFAR-10}  & PGD-AT            & 20.29 ± 0.59 & 6.08 ± 1.71   \\ 
                           & PGD-AT + \textbf{DeMem}    & \textbf{12.39 ± 1.09} & \textbf{1.89 ± 0.79}   \\ 
                           & TRADES         & 12.07 ± 0.69 & 3.60 ± 0.78   \\  
                           & TRADES + \textbf{DeMem} & \textbf{8.47 ± 1.09}   & \textbf{2.56 ± 0.73}   \\ 
\end{tabular}%
}
\label{table:privacy}
\end{table}

\begin{figure}[htb]
\centering
\includegraphics[width=8.8cm]{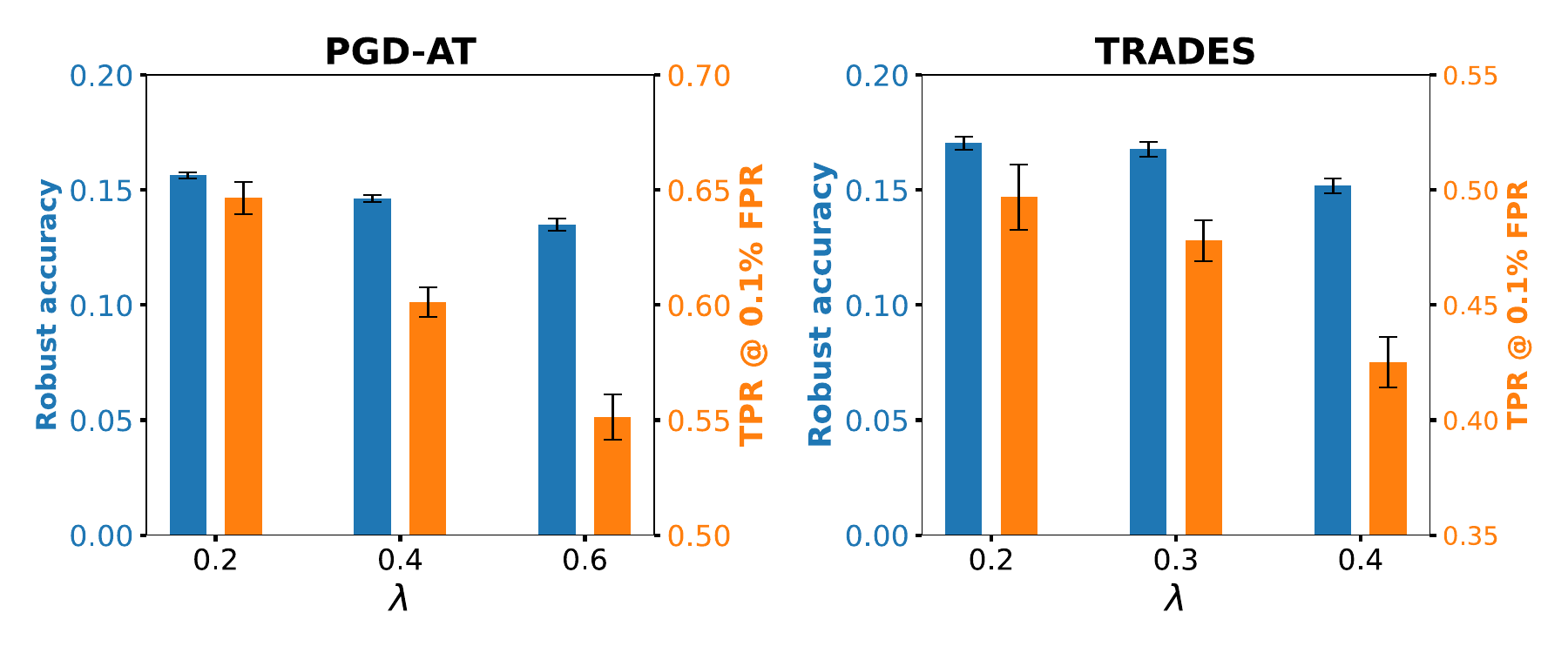}
\caption{Robust accuracy and membership privacy leakage of the proposed DeMem when applying to PGD-AT and TRADES under different $\lambda$ on CIFAR-100.}
\label{fig:diff_lambda_pgd}
\end{figure}

\textbf{Influence of dememorization penalty $\lambda$}: We investigate the effect of the dememorization parameter $\lambda$ on performance. Intuitively, increasing $\lambda$ enhances privacy by restricting the model’s ability to fit high-risk samples, which is confirmed in Fig. \ref{fig:diff_lambda_pgd} by a notable reduction in privacy leakage. Although robustness slightly declines, the drop is much smaller than the privacy improvement.

Although adjusting $\lambda$ may seem to involve a trade-off between privacy and robustness, the resulting performance degradation is negligible compared to the substantial privacy gains. This demonstrates that DeMem is stable across practical parameter settings, offering a flexible and reliable way to enhance privacy while maintaining robustness.

\textbf{Influence of adversarial perturbation magnitude}: 
Since the perturbation norm $\epsilon$ is a key parameter in adversarial training, in Fig. \ref{fig:diff_lambda_pgd}, we compare the attack results of different $\epsilon$ using PGD-AT on CIFAR-100.  Notably, after applying our proposed DeMem method, there is a significant reduction in privacy leakage across various adversarial perturbation levels, while robustness accuracy remains largely unchanged. This confirms the effectiveness and broad applicability of DeMem in enhancing privacy without sacrificing robustness across various adversarial perturbations.

\begin{figure}[htb]
\centering
\includegraphics[width=8.8cm]{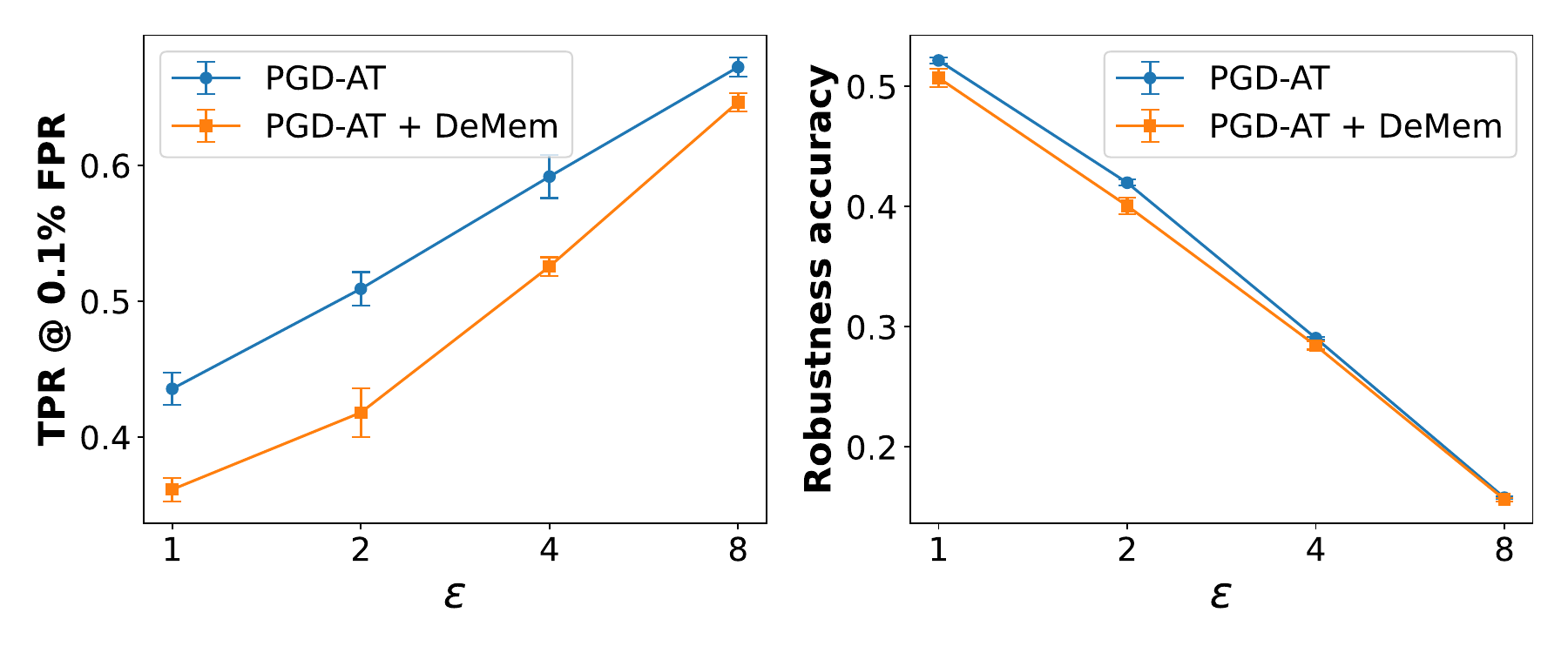}
\caption{LiRA attack success rates and robustness accuracy of PGD-AT and PGD-AT + DeMem on CIFAR-100 under different $\varepsilon$ (Adversarial samples generated with corresponds $\varepsilon$ and 20 iteration steps), the dememorization $\lambda$ = 0.2.}
\label{fig:diff_eps}
\end{figure}

\section{Conclusion and future work}
\label{sec:copyright}
In this paper, we propose a novel privacy-enhanced adversarial training method. We begin by conducting a nuanced analysis to identify which samples contribute to the significant accuracy drop after applying differential privacy (DP), using the memorization score of individual samples. Our analysis reveals that the degradation is primarily caused by samples with low privacy risks being disproportionately affected by DP. This observation inspired us to propose DeMem, a sample-wise dememorization approach designed to mitigate this issue. Through comprehensive empirical validation, we show that DeMem effectively reduces privacy leakage while maintaining comparable robustness, achieving a better balance between privacy and performance in robust models.

These promising results open avenues for future work, including exploring adaptive dememorization techniques based on dataset characteristics and extending evaluations to diverse datasets and attack scenarios.

\vfill\pagebreak

\label{sec:refs}
\bibliographystyle{IEEEbib}
\begingroup
\small
\setlength{\itemsep}{0pt} 
\setlength{\parskip}{0pt} 

\bibliography{icme2025references}
\endgroup
\end{document}